\pdfoutput=1

\documentclass[11pt]{article}

\usepackage{acl}
\usepackage{CJKutf8}

\usepackage{times}
\usepackage{latexsym}

\usepackage[T1]{fontenc}

\usepackage[utf8]{inputenc}

\usepackage{microtype}

\usepackage{booktabs,tabularx}
\usepackage{todonotes}
\usepackage[capitalise,noabbrev]{cleveref}

\def\gptqe{GEMBA}

\def\gembada{\emph{GEMBA-DA}}
\def\gembasqm{\emph{GEMBA-SQM}}
\def\gembastars{\emph{GEMBA-stars}}
\def\gembaclasses{\emph{GEMBA-classes}}

\def\parcite#1{\citep{#1}} 
\def\perscite#1{\citet{#1}} 
\def\inparcite#1{\citealp{#1}} 

\usepackage{multirow}
\usepackage{fancyvrb}

\title{Large Language Models Are State-of-the-Art \\ Evaluators of Translation Quality}

\author{Tom Kocmi \and Christian Federmann \\
  Microsoft,
  One Microsoft Way,
  Redmond, WA-98052, USA \\
  \texttt{\{tomkocmi,chrife\}@microsoft.com}} 

\begin{document}
\begin{CJK}{UTF8}{min}

\maketitle

\begin{abstract}
We describe \gptqe, a GPT-based metric for assessment of translation quality, which works both with a reference translation and without. In our evaluation, we focus on zero-shot prompting, comparing four prompt variants in two modes, based on the availability of the reference. We investigate nine versions of GPT models, including ChatGPT and GPT-4. We show that our method for translation quality assessment only works with GPT~3.5 and larger models.
Comparing to results from WMT22’s Metrics shared task, our method achieves state-of-the-art accuracy in both modes when compared to MQM-based human labels. Our results are valid on the system level for all three WMT22 Metrics shared task language pairs, namely English into German, English into Russian, and Chinese into English. This provides a first glimpse into the usefulness of pre-trained, generative large language models for quality assessment of translations. 
We publicly release all our code and prompt templates used for the experiments described in this work, as well as all corresponding scoring results, to allow for external validation and reproducibility.\footnote{\scriptsize\url{https://github.com/MicrosoftTranslator/GEMBA}}
\end{abstract}

\section{Introduction}
One of the interesting properties of large language models (LLMs) such as GPT~\cite{GPT} is their (implicit) support for multilingual Q\&A. Prompting the model in the right way allows us to translate text between languages \parcite{vilar2022prompting}. This is surprising as GPT has not been fine-tuned for the translation task. 

\perscite{hendy2023good} show that GPT-enabled translation achieves high quality when applied for the translation of high-resource languages, but still lacks in terms of translation quality for under-represented languages.
Building on this finding---\emph{if the model can translate, it may be able to differentiate good from bad translations}---we apply GPT for the task of translation quality assessment. 

In the remainder of this paper, inspired by recent progress on generative, pre-trained large language models (LLMs), we explore how these models can be applied for automated assessment of translation quality. The primary query for this study centers around the question: \emph{Can LLMs be used for effective quality assessment of translations?}

We propose \gptqe, which stands for \emph{GPT Estimation Metric Based Assessment}. The metric evaluates each segment translation in isolation and then averages across all obtained scores for the final, system-level score.

We define and evaluate several prompt variants for zero-shot assessment of translation quality in two modes, either with a human reference translation, as a quality metric, or without one, as a quality estimation task.

We design the main prompts based on the DA+SQM template used for human assessment of translation quality as implemented in the Appraise framework~\cite{federmann-2018-appraise} for WMT22~\cite{kocmi-EtAl:2022:WMT}, building on previous work conducted by \citet{freitag-etal-2021-experts}. 

To the best of our knowledge, our research represents the pioneering effort in exploring the utilization of large language models (LLMs) for the purpose of quality assessment. Subsequent to the publishing of our findings, \perscite{Lu2023EAPrompt} independently published a related report, corroborating the high performance of LLMs.

The main contributions of this paper are:

\begin{itemize}
    \item[-] We demonstrate state-of-the-art capabilities of GPT-based translation quality assessment on the latest WMT22 metrics evaluation data (on the system level);
    \item[-] We experiment with four prompt templates, showing that the least constrained template achieves the best performance;
    \item[-] We evaluate nine different models of GPT, showing that only GPT~3.5 and larger models are capable of translation quality assessment;
    \item[-] We show that GEMBA with GPT-4 model is only slightly behind on segment-level scores to the best-performing metrics.
\end{itemize}

\begin{figure*}[htb]
\input{prompt_da}
\caption{The best-performing prompt based on Direct Assessment expecting a score between 0--100. Template \textbf{portions in bold face} are used only when a human reference translation is available.}
\label{fig:prompt_example}
\end{figure*}

\section{The GEMBA Metric}
To assess translation quality via prompting an LLM, the following parameters are needed: 

\begin{itemize}
\item[-] prompt variant (from a pre-defined set)\\[-2em]
\item[-] source language name, e.g., ``Chinese''\\[-2em]
\item[-] target language name, e.g., ``English''\\[-2em]
\item[-] source segments $src_{1..N}$\\[-2em]
\item[-] candidate translations $hyp_{1..N}$\\[-2em]
\item[-] optionally, reference translations $ref_{1..N}$
\end{itemize}
We generate a GPT request for every segment, querying as individual zero-shot problems, and then aggregate results. For this initial proof of concept, we leave improvements such as few-shot queries or document-level context to future work.

\subsection{Prompt variants}

We experiment with four distinct prompt types: modeling two scoring and two classification tasks. For the scoring tasks, first, one based on \textbf{direct assessment} (\gembada), second, another based on recent research efforts on \textbf{scalar quality metrics} (\gembasqm).\footnote{Although names are based on existing techniques for human assessment, they do not match perfectly.} 
As scoring translation quality may be an unnatural task for an LLM, we also design two classification tasks. The first is based on \textbf{one-to-five stars ranking} (\gembastars), which is a style often used when users are asked to review various services or products. The second prompt asks the LLM to label translation quality as one of five discrete \textbf{quality classes} (\gembaclasses).

For each of these four prompt types, we experiment with two modes that differ with respect to the wording of the corresponding query templates which either have access to a human reference or not. 
As an example, we show the \gembada{} prompt in \cref{fig:prompt_example}. Based on token count, this is the least constrained prompt template that we experiment with. The complete set of prompt templates is available in \cref{sec:prompttemplates}.
For naming convention, we mark quality estimation metrics (without reference) with the suffix "[noref]".

\subsection{Scoring process}
The expected scores are in $[0, 100]$ for \gembada{} and \gembasqm{} prompts, same as for human assessment \parcite{graham-etal-2013-continuous}; for \gembastars{} the output ranges from $[1, 5]$ and \gembaclasses{} assigns one of five class labels.

We average segment-level scores to obtain system-level scores. For the \gembaclasses{} metric variant, we assign classes a numerical value $[0-4]$, based on the label, before averaging.

Depending on the GPT model we query, sometimes answers are returned outside these ranges, as text. When we observe such an \emph{invalid} answer, we add randomness and sample more responses, selecting the first answer matching the output range as the final result.

\subsection{GPT models}

We experiment with seven GPT models---\emph{ranging from GPT~2 up to the latest GPT-4 model}---that are described in \cref{tab:gpt_models}.\footnote{\scriptsize\url{https://learn.microsoft.com/en-us/azure/cognitive-services/openai/concepts/models} and \scriptsize\url{https://platform.openai.com/docs/model-index-for-researchers}}
We use the GPT-4 model as the default model for most experiments and compare the performance of other models in \cref{sec:various_models}.
Specifically, we use these models with brief description:

\begin{description}
\item[GPT~2] We use models provided by \citet{radford2019language}, assessing if GPT~2 may be useful for quality assessment---\emph{we find that it is not};\\[-7mm]

\item[Ada] GPT~3. Max request size of 2,048 tokens and training data up to October 2019 \parcite{brown2020language};\\[-7mm]

\item[Babbage] GPT~3. More capable than Ada \parcite{brown2020language};\\[-7mm]

\item[Curie] GPT~3. More capable than Babbage \parcite{brown2020language};\\[-7mm]

\item[Davinci-002] GPT~3.5. Max request size of 4,000 tokens and training data up to June 2021. Uses \texttt{\small FeedME} training;\\[-7mm]

\item[ChatGPT] Improved GPT~3.5 model, fine-tuned using Reinforcement Learning from Human Feedback (RLHF);\\[-7mm]

\item[Davinci-003] GPT~3.5.1. Uses \texttt{\small PPO} training;\\[-7mm]

\item[GPT-3.5-turbo] Davinci-003 model optimized for speed;\\[-7mm]

\item[GPT-4] there is only limited information about GPT-4, see \perscite{openai2023gpt4}.
\end{description}

GPT~3 models are based on \citet{InstructGPT}. The models are sorted based on their estimated power or date of release. We acknowledge that OpenAI has not released detailed information about the architecture and training data behind given models. Most importantly, OpenAI claims that models have been trained with data up until September 2021. It is important as we use testsets prepared and released by December 2022.

\begin{table}[t]
\centering
\scriptsize
\begin{tabular}{llll}
\toprule
Model name & Abbrev. & Model used \\
\midrule
GPT-2       & ---   & \citet{radford2019language}\\
Ada         & ---   & \texttt{text-ada-001}\\
Babbage     & Bab   & \texttt{text-babbage-001}  \\
Curie       & Curie & \texttt{text-curie-001}  \\
Davinci-002 & Dav2  & \texttt{text-davinci-002} \\
ChatGPT     & Chat  & \texttt{text-chat-davinci-002}\\
Davinci-003 & Dav3  & \texttt{text-davinci-003}\\
GPT-3.5-turbo & Turbo  & \texttt{gpt-3.5-turbo}\\
GPT-4 & GPT4  & \texttt{gpt-4}\\
\bottomrule
\end{tabular}
\caption{Details of all models used in this work. Models are sorted from oldest to newest which also reflects their respective power. GPT~2 and Ada do not work.} 
\label{tab:gpt_models}
\end{table}

\section{Experiments}
To measure the performance of the proposed GEMBA metric, we follow the methodology and use test data provided by the WMT22 Metrics shared task \parcite{freitag-EtAl:2022:WMT} which hosts an annual evaluation of automatic metrics, benchmarking them against human gold labels. Effectively, we compare GEMBA against the best-performing automatic metrics: COMET \parcite{rei-etal-2020-comet, rei-EtAl:2022:WMT1}, BLEURT \parcite{sellam-etal-2020-bleurt}, or the non-public winner MetricX XXL.

\subsection{Test set}
We use the MQM 2022 test set which contains human judgments for the following three translation directions: English into German, English into Russian, and Chinese into English. The test set contains a total of 54 machine translation system outputs or human translations. It contains a total of 106k segments. Translation systems are mainly from participants of the WMT22 General MT shared task \parcite{kocmi-EtAl:2022:WMT}. 

The source segments and human reference translations for each language pair contain around 2,000 sentences from four different texts domains: news, social, conversational, and e-commerce. 
The gold standard for scoring translation quality is based on human MQM ratings, annotated by professionals who mark individual errors in each translation, as described in \perscite{freitag-etal-2021-experts}.

\subsection{Evaluation methods}
To determine how well automatic metrics correlate with humans, we measure system-level, pairwise accuracy (\emph{accuracy}, \inparcite{kocmi-etal-2021-ship}). For segment-level evaluation, we use Kendall's Tau ($\tau$, \inparcite{freitag-etal-2022-results}). 

Here, accuracy is defined as the number of system pairs ranked correctly by the metric with respect to the human ranking divided by the total number of system pair comparisons.

Formally:

\begin{center}
\footnotesize
\[\mbox{Accuracy} = \frac{|\mbox{sign}(\mbox{metric} \Delta)==\mbox{sign}(\mbox{human} \Delta)|}{|\mbox{all system pairs}|}\]
\end{center}

The variant of Kendall's Tau used for metric evaluation changed over the years. Initially, \perscite{callison-burch-etal-2011-findings} proposed to use Kendall's Tau-a ignoring human rankings that tied, while penalising ties in automatic metrics.

\begin{center}
\footnotesize
\[\tau = \frac{|\mbox{Concordant}| - |\mbox{Discordant}|}{|\mbox{Concordant}| + |\mbox{Discordant}|}\]
\end{center}

where Concordant is the set of all human segment comparisons for which a given metric suggests the same order of systems and Discordant is the set of all human comparisons for which a given metric disagrees. 

\begin{center}
\footnotesize
\begin{tabular}{ll|ccc}
 & &   \multicolumn{3}{c}{Metric} \\
 & &     $s_1 < s_2$ & $s_1 = s_2$ & $s_1 > s_2$ \\
\midrule
\multirow{3}{*}{\rotatebox[origin=c]{90}{Human}} & $s_1 < s_2$ & Conc & Disc & Disc\\
& $s_1 = s_2$                        & -- & -- & -- \\
& $s_1 > s_2$                     & Disc & Disc & Conc \\
\end{tabular}
\vspace{1em}
\end{center}

This definition was later updated by \perscite{machacek-bojar-2014-results}, who handle ties as a separate group in contrast to Concordant and Discordant. 
Metrics shared tasks \perscite{mathur-etal-2020-results} and \perscite{freitag-etal-2021-results} changed this back to the 2011 version. Last year, \perscite{freitag-etal-2022-results} changed it to Kendall's Tau-b, which makes adjustments for ties, we use the latest definition in our experiments.
Overall, ties in automatic metrics are rare for non-identical translations but are an issue when a method outputs only a discrete set of scores (as in our case).
Additionally, Kendall's Tau is susceptible to noise in gold pairwise rankings \cite{freitag-etal-2022-results}. 




%

%
We reproduced all scores reported in the WMT22 Metrics shared task findings paper with the official WMT22 script.\footnote{\scriptsize\url{https://github.com/google-research/mt-metrics-eval}} Reported scores match Table 11 of the WMT22 metrics findings paper \parcite{freitag-EtAl:2022:WMT}.

\section{Results}
We investigate GEMBA's performance for two modes: with a reference translation and without reference translation (in a quality estimation setting).
\cref{tab:allmain} reports pairwise accuracy on the system level, comparing \gembada{} against the best-performing metrics from the WMT22 Metrics shared task~\parcite{freitag-EtAl:2022:WMT}. We use GPT-4 as the main model and \gembada{} as the main style for some experiments.

\begin{table}[t]
\footnotesize
\centering
\begin{tabular}{ll}
\toprule
                       Metric & Accuracy \\
\midrule
       \textbf{GEMBA-GPT4-DA} &   89.8\% \\
\textbf{GEMBA-GPT4-DA[noref]} &   87.6\% \\
                  MetricX XXL &   85.0\% \\
                    BLEURT-20 &   84.7\% \\
                     COMET-22 &   83.9\% \\
                     COMET-20 &   83.6\% \\
                        UniTE &   82.8\% \\
                  MS-COMET-22 &   82.8\% \\
                       MATESE &   81.0\% \\
                       YiSi-1 &   79.2\% \\
             COMETKiwi[noref] &   78.8\% \\
              COMET-QE[noref] &   78.1\% \\
                    BERTScore &   77.4\% \\
             UniTE-src[noref] &   75.9\% \\
        MS-COMET-QE-22[noref] &   75.5\% \\
             MATESE-QE[noref] &   74.8\% \\
                   f200spBLEU &   74.1\% \\
                         chrF &   73.4\% \\
                         BLEU &   70.8\% \\
\bottomrule
\end{tabular}

\caption{Results for the system-level pairwise accuracy compared to the current automatic metric. Metrics marked as ``[noref]'' do not use a reference translation.}
\label{tab:allmain}
\end{table}

\subsection{Reference-based}
The results in \cref{tab:allmain} show that our reference-based \textbf{GEMBA-GPT4-DA} metric sets a new state of the art. It outperforms all of the other reference-based metrics from the WMT22 Metrics shared task.
The observed level of metric performance is unexpected, especially considering that human labels used as a gold standard in itself are noisy and therefore an accuracy of 100\% is impossible to obtain for an automatic metric.

\subsection{Quality estimation}
\cref{tab:allmain} shows that our reference-less metric \textbf{GEMBA-GPT4-DA[noref]} achieves the highest performance for the quality estimation mode, and strongly outperforms all of the other reference-less metrics. Moreover, it also outperforms all of the other reference-based metrics, performing only slightly worse than \textbf{GEMBA-GPT4-DA}. Again, the observed level of assessment quality is unexpectedly high, highlighting the potential of using LLMs for translation quality assessment tasks.

\begin{table*}[t]
\small
\centering
\begin{tabular}{rrrrrrrr}
\toprule
               &    Bab &  Curie &            Dav2 &            Chat &            Dav3 &           Turbo &            GPT4 \\
\midrule
            DA & 39.1\% & 54.4\% & \textbf{85.8\%} &          81.0\% & \textbf{88.0\%} & \textbf{86.5\%} & \textbf{89.8\%} \\
     DA[noref] & 55.8\% & 51.8\% &          83.9\% &          82.1\% & \textbf{86.1\%} & \textbf{86.9\%} & \textbf{87.6\%} \\
           SQM & 51.8\% & 40.5\% & \textbf{85.8\%} & \textbf{85.0\%} & \textbf{85.4\%} & \textbf{87.2\%} & \textbf{88.7\%} \\
    SQM[noref] & 51.1\% & 41.6\% &          82.8\% &          81.0\% &          82.5\% & \textbf{87.6\%} & \textbf{89.1\%} \\
         Stars & 48.2\% & 37.2\% & \textbf{88.3\%} & \textbf{85.0\%} & \textbf{85.8\%} & \textbf{89.4\%} & \textbf{91.2\%} \\
  Stars[noref] & 58.4\% & 54.7\% &          79.6\% &          83.6\% &          83.2\% &          84.3\% & \textbf{89.1\%} \\
       Classes & 47.4\% & 43.4\% &          79.6\% & \textbf{87.2\%} & \textbf{85.4\%} &          82.5\% & \textbf{89.1\%} \\
Classes[noref] & 35.0\% & 61.7\% &          78.1\% &          83.6\% &          78.8\% &          62.0\% & \textbf{91.2\%} \\
\bottomrule
\end{tabular}

\caption{Accuracy of the system-level pairwise accuracy for quality estimation methods for most combinations of prompts and different GPT models. The evaluation is based on three language pairs and MQM human labels. All results higher than the WMT22 winner of Metrics shared task MetricX XXL are bolded.}
\label{tab:all_models}
\end{table*}

\subsection{Comparison of GPT models}
\label{sec:various_models}

We compare the performance of various GPT versions as an automatic metric. \cref{tab:all_models} shows results for all models we have experimented with and all prompt variants tested.

We do not show results for GPT-2 or Ada models. Neither of those have produced replies in the specific scoring range and neither seemed to be producing any meaningful replies. We list a couple of their answers in \cref{sec:adareplies}. Based on our experiments, we conclude that they are not powerful enough to understand the zero-shot prompts.

By contrast, Babbage and Curie models appear to understand what type of answer they should produce, but the quality of their scores seems to be close to random guessing. Thus, both Babbage and Curie are useless for translation quality assessment.

The main performance jump occurs for GPT 3.5 and larger models, i.e., Davinci-002, ChatGPT, Davinci-003, Turbo, and GPT-4. Each of them achieves highly competitive results for all of the prompt variants we have tested. Interestingly, ChatGPT in DA style appears to have the lowest quality among those models. In addition, ChatGPT and Turbo frequently reply with a score followed by an explanation of why it has assigned that score.
One possible reason may be in the form of the prompt, which wasn't modified to instruct ChatGPT not to generate an explanation.

Unsurprisingly, the best performance is obtained by the most powerful LLM, GPT-4. Moreover, we can see that over time, each generation of models is slightly better. This confirms the findings of \perscite{hendy2023good} who demonstrated superior translation capabilities with Davinci-003 over all other previous GPT variants.

\subsection{Segment-level performance}

All previous results are reported on the system level. We also investigate how well the GEMBA metric performs on the segment level, with respect to the human gold annotations. We present Kendall's Tau results for each language pair separately in \cref{tab:seglevel} for GPT-4 and Davinci-003 (results for all metrics are in \cref{tab:all_results}).

GPT-4 models are slightly behind the top-performing metrics but continue to have a high correlation with human judgment. On the other hand, quality estimation \textbf{GEMBA-Dav3-DA [noref]} has significantly lower segment-level performance in contrast to other top-performing metrics. 

The lower performance of a segment-level correlation could be attributed to Kendall's Tau, which penalizes ties. Our metric in contrast to other automatic metrics returns a discrete value between 0--100. There is a high probability that two translations will obtain an equal score.

\begin{table}[t]
\scriptsize
\centering
\begin{tabular}{lllll}
\toprule
              Metric &    Acc &         en-de &         en-ru &         zh-en \\
\midrule
       GEMBA-GPT4-DA & 89.8\% &          0.36 &          0.36 &          0.38 \\
       GEMBA-Dav3-DA & 88.0\% &          0.31 &          0.33 &          0.37 \\
GEMBA-GPT4-DA[noref] & 87.6\% &          0.31 &          0.40 &          0.41 \\
GEMBA-Dav3-DA[noref] & 86.1\% &          0.18 &          0.26 &          0.29 \\
         MetricX XXL & 85.0\% &          0.36 & \textbf{0.42} & \textbf{0.43} \\
           BLEURT-20 & 84.7\% &          0.34 &          0.36 &          0.36 \\
            COMET-22 & 83.9\% & \textbf{0.37} &          0.40 & \textbf{0.43} \\
               UniTE & 82.8\% & \textbf{0.37} &          0.38 &          0.36 \\
    COMETKiwi[noref] & 78.8\% &          0.29 &          0.36 &          0.36 \\
     COMET-QE[noref] & 78.1\% &          0.28 &          0.34 &          0.36 \\
                chrF & 73.4\% &          0.21 &          0.17 &          0.15 \\
                BLEU & 70.8\% &          0.17 &          0.14 &          0.14 \\
\bottomrule
\end{tabular}

\caption{Kendall's Tau ($\tau$) segment-level evaluation. Full results are in \cref{tab:all_results}.}
\label{tab:seglevel}
\end{table}

In order to investigate this further, we collect all answers across all systems and all three language pairs and then calculate the frequency of each distinct answer value.

We can notice several interesting observations in \cref{tab:score_distribution}. The DA reference-based prompt generates mostly multiples of five. Over three-quarters of all scores are either score 80, 95, or 100. This could reflect the actual quality of the system translations as the underlying systems are provably high-quality. This is also a finding of \perscite{freitag-EtAl:2022:WMT} that many metrics fall into the same significance cluster.

When we investigate the ``DA[noref]'', we notice that 60.5\% of all scores are of value "95". Despite this fact, the metric still manages to differentiate the systems from each other and outperform all other quality estimation metrics on the system level. This is contributed to the fact that better-performing systems obtain more segments with a score 95 than worse-performing systems, therefore getting a lower average score. We should note, that there are no system-level ties.

We conjecture that frequent segment-level ties and the discrete scale thus may contribute to the lower Kendall's Tau segment-level performance.

\begin{table}[t]
\scriptsize
\centering
\begin{tabular}{rrrrr}
\toprule
Answers &    DA & DA[noref] &   SQM & SQM[noref] \\
\midrule
      0 &  0.1\% &      0.1\% &  0.1\% &       0.1\% \\
      5 &  0.0\% &      0.0\% &  0.0\% &       0.0\% \\
     10 &  0.0\% &      0.0\% &  0.0\% &       0.1\% \\
     15 &   --- &       --- &  0.0\% &       0.0\% \\
     20 &  0.2\% &      0.3\% &  0.2\% &       0.3\% \\
     25 &   --- &       --- &  0.0\% &        --- \\
     30 &  0.1\% &      0.2\% &  0.1\% &       0.1\% \\
     35 &   --- &       --- &  0.0\% &        --- \\
     40 &  0.5\% &      0.6\% &  0.5\% &       0.6\% \\
     45 &  0.0\% &      0.0\% &  0.0\% &       0.0\% \\
     50 &  0.0\% &      0.1\% &  0.1\% &       0.0\% \\
     55 &   --- &       --- &  0.0\% &        --- \\
     60 &  2.1\% &      2.3\% &  2.0\% &       2.1\% \\
     65 &   --- &      0.0\% &  0.0\% &       0.0\% \\
     70 &  1.3\% &      0.4\% &  1.9\% &       0.6\% \\
     75 &  0.5\% &      1.0\% &  0.7\% &       0.7\% \\
     80 &  6.3\% &      4.5\% &  7.0\% &       5.7\% \\
     85 &  4.4\% &      2.7\% &  6.0\% &       2.9\% \\
     87 &   --- &       --- &  0.0\% &        --- \\
     88 &   --- &       --- &  0.0\% &        --- \\
     90 & 21.3\% &     13.0\% & 27.6\% &      14.5\% \\
     92 &   --- &       --- &  0.0\% &        --- \\
     93 &   --- &       --- &  0.0\% &        --- \\
     94 &   --- &       --- &  0.0\% &        --- \\
     95 & 53.3\% &     60.6\% & 44.6\% &      49.4\% \\
     98 &  0.8\% &      0.0\% &  0.4\% &       0.0\% \\
     99 &  0.4\% &       --- &  0.2\% &        --- \\
    100 &  8.6\% &     14.1\% &  8.5\% &      22.8\% \\
\bottomrule
\end{tabular}

\caption{Distribution of all distinct segment-level score values for MQM 2022 for model GPT-4.}
\label{tab:score_distribution}
\end{table}

\subsection{Failure rate}

As we described earlier, LLMs may answer with an invalid answer, for example with a textual answer instead of a score, mostly explaining its decision. When such a situation happens, we iteratively increase the temperature---\emph{adding randomness to the model}---and take the first answer matching the expected score output range.

This adds non-determinism to our evaluation, therefore we investigate how frequently this phenomenon happens. \cref{tab:failrate} shows the number of invalid answers. For almost all combinations of models and prompts, except of SQM-style, LLMs understand the prompt and provide answers in a valid range with less than 1\% of the answers being invalid.\footnote{Roughly 1,000 answers equal to 1\% of the total volume.} This has a minimal effect on the final system-level score and therefore, we conclude that the metric is mostly deterministic.

Additionally, we confirm that a temperature equal to zero always returns the same answer, which we evaluated by re-running GEMBA-Dav2-DA[noref].

\begin{table*}[t]
\footnotesize
\centering
\begin{tabular}{lrrrrrr}
\toprule
               &    Bab &  Curie &  Dav2 &  Chat &  Dav3 &  GPT4 \\
\midrule
            DA &    750 &  8,048 &     7 &   565 &     0 &     0 \\
     DA[noref] &    146 &    862 &     0 &   935 &    53 &     0 \\
           SQM & 89,599 &    129 & 4,827 &    45 & 1,279 &   --- \\
    SQM[noref] & 15,577 & 95,131 & 1,763 &    59 &     1 &     0 \\
         Stars & 18,074 &    --- &   135 & 1,064 &    58 &   --- \\
  Stars[noref] &    --- & 86,593 &   135 & 1,924 &     1 &     0 \\
       Classes &     74 &     12 &     0 &    10 &     0 &   --- \\
Classes[noref] &    115 &     15 &     0 &    12 &     0 &   --- \\
\bottomrule
\end{tabular}

\caption{Number of invalid answers (full set size 106,758) that needed to be re-prompted with added randomness. The evaluation of ChatGPT and parts of GPT-4 were excluded due to their late integration and changes in our codebase.}
\label{tab:failrate}
\end{table*}

Processing answers is straightforward as it is usually a stand-alone number. In some occasions, LLMs give a numerical score and continue with a textual explanation, for such cases, we parse only the first number.
A more complex approach needs to be taken for \gembastars{} prompts where the model provides different answers which we parse separately. Here are some examples of two-star answers: "2", "two", "**", "★★", "two stars", or "2 stars". For non-English target languages the answer may be produced in the target language, e.g., "一颗星", or "五".
We have not observed attempts to translate output for other prompts.

\section{Conclusion}
We have presented our work on GEMBA, a GPT-based estimation metric-based assessment method. Comparing our metrics to other automated metrics from the WMT22 Metrics shared task we report state-of-the-art performance on the MQM 2022 test set across three language pairs: English to German, English to Russian, and Chinese to English.

We intend to continue research on the application of GPT models for quality assessment. Further research will focus on the switch to few-shot (as opposed to our current zero-shot methodology) as well as model fine-tuning. Both of which promise to increase GEMBA accuracy. Furthermore, modifying prompts to support MQM error-based evaluation or post-editing efforts may lead to further improvements.



GPT-enhanced evaluation metrics may allow us to make progress with respect to document-level evaluation (due to their ability to use much larger context windows). This could be beneficial as there is little research into document-level metrics \parcite{vernikos-EtAl:2022:WMT}.

\section*{Limitations}
While preliminary results indicate that the GEMBA metric performs very well when compared to other automated metrics evaluated as part of the WMT22 Metrics shared task, it is important to note that these results are based on human labels for {\emph{only three language pairs}}. We expect that the metrics performance may suffer for other language pairs, mainly under-resourced languages similar to \perscite{hendy2023good} who show low translation quality for such languages. In addition, GEMBA's state-of-the-art performance only holds for the system level, while segment-level scores still have room for improvement.
Reported results are indicative of the potential performance LLMs could achieve for the translation quality assessment task in the long run. However, more analysis is needed before using it as the main tool for deciding translation quality.

An additional limitation to consider in this study is the inability to definitively ascertain that the evaluation data have not been included in OpenAI's training dataset. Nevertheless, the available evidence strongly indicates that this is unlikely. OpenAI claims that their data compilation only extends up to September 2021, while the test set employed in this research was generated during the second half of 2022 and made publicly available in December 2022. Our initial positive results using the Davinci-002 model were obtained in early February, which presents a narrow timeframe for OpenAI to incorporate and process the evaluation data. Furthermore, the test set is not readily accessible in plaintext format, necessitating pre-processing prior to utilization in training.

\section*{Acknowledgments}
This work would not have been possible without the help and support from our friend and colleague, Olivier Nano, who provided access to GPT models via Microsoft Azure -- {\emph{Merci beaucoup, Olivier!}} 
The authors are also grateful to Matt Post, Vikas Raunak, Shabnam Sadegharmaki, and the Microsoft Translator research team for fruitful discussions and helpful feedback.

\bibliography{anthology,custom}

\clearpage
\appendix
\clearpage
\onecolumn
\section{Appendix: Prompt Templates}
\label{sec:prompttemplates}
Below we provide our prompt templates which we use for the experiments described in this paper. Template \textbf{portions in bold face} are used only when a human reference translation is available.

\subsection{DA: Direct Assessment}
Output scores range from $0-100$.\\
 {\footnotesize
    \begin{Verbatim}[commandchars=+\[\]]
    Score the following translation from {source_lang} to {target_lang} +textbf[with respect to]
    +textbf[the human reference] on a continuous scale from 0 to 100, where a score of zero means
    "no meaning preserved" and score of one hundred means "perfect meaning and grammar".

    {source_lang} source: "{source_seg}"
    +textbf[{target_lang} human reference: {reference_seg}]
    {target_lang} translation: "{target_seg}"
    Score:
    \end{Verbatim}
}
\subsection{SQM: Scalar Quality Metrics}
Output scores range from $0-100$.\\
 {\footnotesize
    \begin{Verbatim}[commandchars=+\[\]]
    Score the following translation from {source_lang} to
    {target_lang} +textbf[with respect to the human reference] on a continuous
    scale from 0 to 100 that starts with "No meaning preserved", goes
    through "Some meaning preserved", then "Most meaning preserved and
    few grammar mistakes", up to "Perfect meaning and grammar".

    {source_lang} source: "{source_seg}"
    +textbf[{target_lang} human reference: "{reference_seg}"]
    {target_lang} translation: "{target_seg}"
    Score (0-100):
    \end{Verbatim}
}
\subsection{Stars: One to Five Stars Ranking}
Output scores range from $1-5$. Special care is taken for answers containing non-numerical answers, such as "Three stars", "****", or "1 star".\\
 {\footnotesize
    \begin{Verbatim}[commandchars=+\[\]]
    Score the following translation from {source_lang} to {target_lang} 
    +textbf[with respect to the human reference] with one to five stars. 
    Where one star means "Nonsense/No meaning preserved", 
    two stars mean "Some meaning preserved, but not understandable", 
    three stars mean "Some meaning preserved and understandable", 
    four stars mean "Most meaning preserved with possibly few grammar mistakes", 
    and five stars mean "Perfect meaning and grammar".
    
    {source_lang} source: "{source_seg}"
    +textbf[{target_lang} human reference: "{reference_seg}"]
    {target_lang} translation: "{target_seg}"
    Stars:
    \end{Verbatim}
}
\subsection{Classes: Quality Class Labels}
Output label one of {\tt "No meaning
    preserved"}, {\tt "Some meaning preserved, but not understandable"}, {\tt "Some meaning preserved and understandable"}, {\tt "Most meaning preserved, minor issues"}, {\tt "Perfect translation"}.\\
 {\footnotesize
    \begin{Verbatim}[commandchars=+\[\]]
    Classify the quality of translation from {source_lang} to {target_lang}
    +textbf[with respect to the human reference] into one of following classes: "No meaning
    preserved", "Some meaning preserved, but not understandable", "Some meaning
    preserved and understandable", "Most meaning preserved, minor issues", "Perfect
    translation".

   {source_lang} source: "{source_seg}"
   +textbf[{target_lang} human reference: "{reference_seg}"]
   {target_lang} translation: "{target_seg}"
   Class:
    \end{Verbatim}
}

\noindent Templates available from: {\small \url{https://github.com/MicrosoftTranslator/GEMBA/gemba/prompt.py}}
\onecolumn
\section{Appendix: Full Results}
\label{sec:fullresults}

Below table lists all GEMBA results we have obtained for this work. Any missing segment-level scores are due to a subset of segments for which we could not obtain a score even after adding randomness.

\begin{table*}[hbt!]
\scriptsize
\centering
\begin{tabular}{lllll}
\toprule
                    Metric & Accuracy &          en-de &          en-ru &          zh-en \\
\midrule
 GEMBA-GPT4-Classes[noref] &   91.2\% &          0.304 &          0.390 &          0.313 \\
          GEMBA-GPT4-Stars &   91.2\% &          0.326 &          0.351 &          0.382 \\
             GEMBA-GPT4-DA &   89.8\% &          0.357 &          0.358 &          0.382 \\
         GEMBA-Turbo-Stars &   89.4\% &          0.259 &          0.223 &          0.265 \\
        GEMBA-GPT4-Classes &   89.1\% &          0.222 &          0.267 &          0.273 \\
   GEMBA-GPT4-Stars[noref] &   89.1\% &          0.308 &          0.366 &          0.404 \\
     GEMBA-GPT4-SQM[noref] &   89.1\% &          0.359 & \textbf{0.432} &          0.416 \\
            GEMBA-GPT4-SQM &   88.7\% & \textbf{0.380} &          0.388 &          0.398 \\
          GEMBA-Dav2-Stars &   88.3\% &          0.225 &          0.282 &          0.183 \\
             GEMBA-Dav3-DA &   88.0\% &          0.306 &          0.332 &          0.371 \\
      GEMBA-GPT4-DA[noref] &   87.6\% &          0.311 &          0.405 &          0.407 \\
    GEMBA-Turbo-SQM[noref] &   87.6\% &          0.259 &          0.309 &          0.291 \\
        GEMBA-Chat-Classes &   87.2\% &          0.220 &          0.270 &          0.259 \\
           GEMBA-Turbo-SQM &   87.2\% &          0.298 &          0.277 &          0.313 \\
     GEMBA-Turbo-DA[noref] &   86.9\% &          0.255 &          0.294 &          0.264 \\
            GEMBA-Turbo-DA &   86.5\% &          0.250 &          0.234 &          0.255 \\
      GEMBA-Dav3-DA[noref] &   86.1\% &          0.180 &          0.258 &          0.289 \\
          GEMBA-Dav3-Stars &   85.8\% &          0.294 &          0.294 &          0.297 \\
            GEMBA-Dav2-SQM &   85.8\% &          0.279 &          0.325 &          0.344 \\
             GEMBA-Dav2-DA &   85.8\% &          0.231 &          0.302 &          0.303 \\
        GEMBA-Dav3-Classes &   85.4\% &          0.235 &          0.289 &          0.251 \\
            GEMBA-Dav3-SQM &   85.4\% &          0.283 &          0.308 &          0.346 \\
               MetricX XXL &   85.0\% &          0.360 &          0.420 &          0.427 \\
          GEMBA-Chat-Stars &   85.0\% &          0.292 &          0.248 &          0.343 \\
            GEMBA-Chat-SQM &   85.0\% &          0.250 &          0.293 &          0.310 \\
                 BLEURT-20 &   84.7\% &          0.344 &          0.359 &          0.361 \\
  GEMBA-Turbo-Stars[noref] &   84.3\% &          0.255 &          0.279 &          0.261 \\
                  COMET-22 &   83.9\% &          0.368 &          0.400 & \textbf{0.428} \\
      GEMBA-Dav2-DA[noref] &   83.9\% &          0.209 &          0.285 &          0.280 \\
                  COMET-20 &   83.6\% &          0.319 &          0.330 &          0.332 \\
 GEMBA-Chat-Classes[noref] &   83.6\% &          0.193 &          0.306 &          0.256 \\
   GEMBA-Chat-Stars[noref] &   83.6\% &          0.209 &          0.323 &          0.356 \\
   GEMBA-Dav3-Stars[noref] &   83.2\% &          0.198 &          0.310 &          0.235 \\
                     UniTE &   82.8\% &          0.369 &          0.378 &          0.357 \\
               MS-COMET-22 &   82.8\% &          0.283 &          0.351 &          0.341 \\
     GEMBA-Dav2-SQM[noref] &   82.8\% &          0.216 &          0.306 &          0.310 \\
     GEMBA-Dav3-SQM[noref] &   82.5\% &          0.218 &          0.328 &          0.268 \\
       GEMBA-Turbo-Classes &   82.5\% &          0.170 &          0.167 &          0.178 \\
      GEMBA-Chat-DA[noref] &   82.1\% &          0.231 &          0.332 &          0.359 \\
                    MATESE &   81.0\% &          0.323 &          0.279 &          0.389 \\
     GEMBA-Chat-SQM[noref] &   81.0\% &          0.224 &          0.320 &          0.284 \\
             GEMBA-Chat-DA &   81.0\% &          0.307 &          0.328 &          0.361 \\
        GEMBA-Dav2-Classes &   79.6\% &          0.173 &          0.260 &          0.184 \\
   GEMBA-Dav2-Stars[noref] &   79.6\% &          0.142 &          0.203 &          0.193 \\
                    YiSi-1 &   79.2\% &          0.235 &          0.227 &          0.296 \\
          COMETKiwi[noref] &   78.8\% &          0.290 &          0.359 &          0.364 \\
 GEMBA-Dav3-Classes[noref] &   78.8\% &          0.176 &          0.271 &          0.172 \\
           COMET-QE[noref] &   78.1\% &          0.281 &          0.341 &          0.365 \\
 GEMBA-Dav2-Classes[noref] &   78.1\% &          0.105 &          0.172 &          0.128 \\
                 BERTScore &   77.4\% &          0.232 &          0.192 &          0.316 \\
          UniTE-src[noref] &   75.9\% &          0.287 &          0.342 &          0.343 \\
     MS-COMET-QE-22[noref] &   75.5\% &          0.233 &          0.305 &          0.287 \\
          MATESE-QE[noref] &   74.8\% &          0.244 &          0.229 &          0.337 \\
                f200spBLEU &   74.1\% &          0.180 &          0.153 &          0.140 \\
                      chrF &   73.4\% &          0.214 &          0.168 &          0.147 \\
                      BLEU &   70.8\% &          0.169 &          0.140 &          0.145 \\
GEMBA-Turbo-Classes[noref] &   62.0\% &         -0.010 &          0.027 &          0.029 \\
GEMBA-Curie-Classes[noref] &   61.7\% &          0.001 &         -0.007 &         -0.053 \\
    GEMBA-Bab-Stars[noref] &   58.4\% &            --- &            --- &            --- \\
       GEMBA-Bab-DA[noref] &   55.8\% &         -0.119 &         -0.011 &            --- \\
  GEMBA-Curie-Stars[noref] &   54.7\% &            --- &            --- &            --- \\
            GEMBA-Curie-DA &   54.4\% &            --- &            --- &            --- \\
     GEMBA-Curie-DA[noref] &   51.8\% &            --- &          0.054 &            --- \\
             GEMBA-Bab-SQM &   51.8\% &            --- &            --- &            --- \\
      GEMBA-Bab-SQM[noref] &   51.1\% &         -0.010 &          0.006 &            --- \\
           GEMBA-Bab-Stars &   48.2\% &            --- &            --- &            --- \\
         GEMBA-Bab-Classes &   47.4\% &         -0.086 &         -0.089 &         -0.066 \\
       GEMBA-Curie-Classes &   43.4\% &         -0.011 &         -0.035 &         -0.035 \\
    GEMBA-Curie-SQM[noref] &   41.6\% &            --- &            --- &            --- \\
           GEMBA-Curie-SQM &   40.5\% &          0.026 &          0.013 &          0.007 \\
              GEMBA-Bab-DA &   39.1\% &         -0.110 &            --- &            --- \\
         GEMBA-Curie-Stars &   37.2\% &            --- &            --- &            --- \\
  GEMBA-Bab-Classes[noref] &   35.0\% &         -0.010 &         -0.033 &          0.067 \\
\bottomrule
\end{tabular}

\label{tab:all_results}
\caption{System-level pairwise accuracy and segment-level Kendall Tau scores.}
\end{table*}
\clearpage
\onecolumn
\section{Appendix: Ada Answers}
\label{sec:adareplies}
We present some of the replies from the Ada model on various prompts in \cref{tab:adareplies} to illustrate the frequent, invalid answers of this model. Similar behaviour was observed when querying GPT~2. Both models failed with respect to our translation quality assessment experiment.

\newcommand{\specialcell}[2][c]{%
  \begin{tabular}[#1]{@{}l@{}}#2\end{tabular}}

\begin{table*}[h]
\centering
\begin{tabular}{ll}
\toprule
DA & \specialcell{0 = no meaning preserved\\1 = perfect meaning and grammar\\2 = excellent meaning\\3 = very good meaning\\4 = excellent\\5 = very good\\6 = excellent\\7 = excellent\\8 = excellent\\9 = excellent\\10 = perfect} \\
\midrule
DA & \specialcell{No meaning preserved: 0 \\ Perfect meaning and grammar: 100} \\
\midrule
DA & \specialcell{0 – It does not exist in my language \\ 1 – It does not exist in my text \\2 – It does not exist in my text \\3 – It exists in my text \\4 – It exists in my text\\5 – It exists in my text\\6 – It exists in my text\\7 – It exists in my text\\8 – It exists in my text\\9 – It exists in my text\\10 – It exists in my text} \\
\midrule
DA & This sentence does not have a score.\\
\midrule
DA[noref]  & \specialcell{The source for this translation is English. The translation is 1 out of 100.} \\
\midrule
DA[noref]  & \specialcell{0 = no meaning preserved\\1 = perfect meaning and grammar\\2 = perfect meaning} \\
\midrule
SQM  & \specialcell{Some meaning preserved\\Most meaning preserved and few grammar mistakes} \\
\midrule
SQM[noref]  & \specialcell{No meaning preserved\\Some meaning preserved\\Most meaning preserved and few grammar mistakes} \\

\bottomrule
\end{tabular}
\caption{Answers by the Ada model for various prompts. We observe that SQM prompts are closer to expected outputs than answers to the corresponding DA prompts. Similar behaviour was observed when querying GPT~2.}
\label{tab:adareplies}
\end{table*}

\end{CJK}
\end{document}